\documentclass[11pt]{article}

\usepackage[preprint]{acl}

\usepackage{times}
\usepackage{latexsym}
\usepackage{amsmath}
\usepackage{amssymb} 
\usepackage{pifont}  
\usepackage{stmaryrd}
\usepackage{hyperref}
\usepackage{mathrsfs}
\usepackage[most]{tcolorbox}
\usepackage[dvipsnames]{xcolor}
\usepackage[T1]{fontenc}
\usepackage[utf8]{inputenc}
\usepackage{float}
\usepackage[most]{tcolorbox}

\usepackage{soul}
\newcommand{\hlc}[2]{{\sethlcolor{#1}\hl{#2}}}

\usepackage{microtype}

\usepackage{inconsolata}

\usepackage{longtable}
\usepackage{array}
\usepackage{ragged2e}

\usepackage{graphicx}
\usepackage{booktabs}
\usepackage{tabularx}
\usepackage{multirow}
\usepackage{textgreek}

\title{Explanation Generation for Contradiction Reconciliation with LLMs}

\author{Jason Chan \hspace{10mm} Zhixue Zhao \hspace{10mm} Robert Gaizauskas \\
        University of Sheffield, UK \\ \{jlychan1, zhixue.zhao, r.gaizauskas\}@sheffield.ac.uk}

\newtcolorbox{promptbox}{
    colback=orange!5!white,    
    colframe=orange!50!black,  
    arc=2mm,                   
    boxrule=0.5pt,             
    left=10pt,                 
    right=10pt,
    top=8pt,
    bottom=8pt,
    fontupper=\ttfamily,       
    breakable                  
}

\begin{document}
\maketitle
\begin{abstract}

Existing NLP work commonly treats contradictions as errors to be resolved by choosing which statements to accept or discard. Yet a key aspect of human reasoning in social and professional settings is the ability to hypothesize explanations that \textit{reconcile} contradictions. For example, ``\textit{Cassie hates coffee}'' and ``\textit{She buys coffee everyday}'' appear contradictory, but are in fact compatible if Cassie was assigned the daily chore of buying coffee for all her coworkers. Despite advances in the reasoning capabilities of large language models (LLMs), their ability to hypothesize such explanations remains largely unexplored. To address this gap, we introduce the task of \textit{reconciliatory explanation generation}, where models must generate explanations that effectively render contradictory statements compatible. We propose a novel method of reusing natural language inference (NLI) datasets, and introduce quality metrics that enable scalable automatic evaluation. Experiments with 13 LLMs in thinking and non-thinking modes reveal models' limited success in this task, with longer reasoning chains negatively associated with explanation success. As a proof of concept, we further adapt our introduced task and metrics for reinforcement-learning training, finding that trained models exhibit improved accuracy in non-thinking mode on two narrative reasoning benchmarks (True Detective and MuSR). Our findings highlight reconciliatory explanation generation as a challenge in LLM reasoning and motivate further work to investigate its potential for improving model capabilities.

\end{abstract}

\begin{figure}
    \centering
    \includegraphics[width=0.97\linewidth]{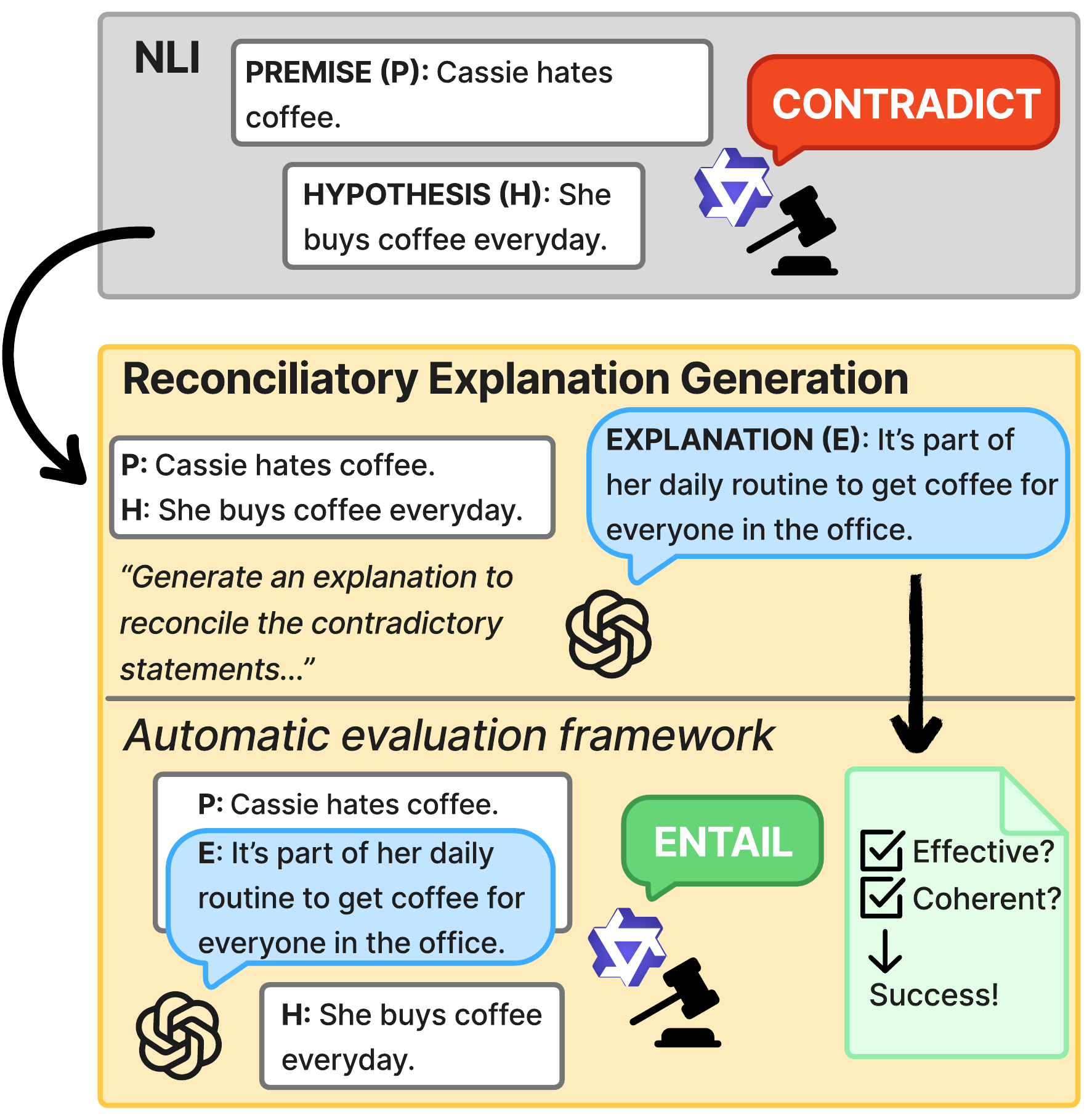}
    \caption{Given a premise (\textit{P}) and hypothesis (\textit{H}) that a judge model deems contradictory, our novel task requires models to generate a successful explanation (\textit{E}) such that \textit{H} is judged as entailed by \textit{P} combined with \textit{E}.}
    \label{fig:introduction_figure}
\end{figure}

\section{Introduction}

\begin{quote}
    ``\textit{How wonderful that we have met with a paradox. Now we have some hope of making progress.}'' — Niels Bohr
\end{quote}

A contradiction is a categorical error in a rigid system (e.g. one relying on classical logic), but it often serves as a crucial trigger for reasoning in human cognition. When humans encounter conflicting information, we rarely resort to simply discarding one observation in favor of another; instead, we engage in abductive reasoning to search for an explanation that reconciles the apparent contradiction between them \cite{doi:10.1080/17470218.2011.592593}. This motivation and ability to \textit{reconcile} rather than simply \textit{arbitrate} between conflicting observations is a hallmark of intelligence that manifests across everyday social interactions, professional and scientific domains \cite{BOHR1928,doi:10.1177/1088868310362982, 10.3389/fpsyg.2022.710815}.

In conversations, when a speaker says something that seems to contradict our existing knowledge (for example, a friend claims to love French cuisine but we know he actively avoids the local French restaurants), we rarely assume that either party must simply be mistaken. Instead, as predicted by Grice's Cooperative Principle \cite{grice1975logic_cooperative_principle} and the Principle of Charity \cite{Quine1960-QUIWO, Davidson1973}, we hypothesize reconciling explanations, i.e., seeking interpretations that preserve coherence given our prior beliefs or existing knowledge (perhaps he only likes home-cooked French food, or he is reluctant to admit that he cannot afford those expensive restaurants). Similarly, in the legal domain, certain US court decisions are based on the canon of ``\textit{harmonious-reading}'' \cite{scalia2012reading}, which requires interpreting provisions in statutes and contracts such that, whenever possible, they are mutually compatible rather than contradictory. 

However, as large language models (LLMs) are increasingly deployed as conversational agents and in scientific and professional settings \cite{majumder2025discoverybench, SINGH2025100128, zheng-etal-2025-automation}, their ability to \textbf{generate explanations to reconcile contradictions between statements or observations} remains largely underexplored. Existing research on how LLMs handle information conflicts predominantly focuses on detection and arbitration: deciding which conflicting statement(s) to accept and which to reject \cite{kazemi2023boardgameqa, wang2024resolving}. While the generation of natural language explanations has been extensively studied in other contexts \cite{10.5555/3327546.3327624,huang2023largelanguagemodelsexplain,10.1007/978-3-031-84595-6_1,chen-etal-2025-rose}, attention is seldom given to assessing explanations for reconciling seemingly contradictory observations. Enabling LLMs to perform such reconciliations is crucial for improving their usefulness in dialogue systems and knowledge-intensive applications where imperfect and conflicting information is common.

To address this gap, we introduce the novel task of \textbf{reconciliatory explanation generation (REG)}, where \textit{a model must generate natural language explanations that effectively render two apparently contradictory statements mutually compatible}. We leverage existing natural language inference (NLI) datasets by recognizing inherent human annotator disagreements as a feature rather than a bug, and introduce new scalable metrics with which we evaluate current LLMs on this novel reasoning task.

Our contributions are as follows:
\begin{enumerate}
    \item We introduce \textit{reconciliatory explanation generation}, a new task to evaluate models' capabilities in reconciling contradictory statements through explanations, an underexplored dimension of LLM reasoning.
    \item We propose a scalable framework of reusing current NLI datasets as data for our task, leveraging existing annotator disagreements whilst not requiring additional human annotations.
    \item We introduce automatic evaluation metrics based on standard three-way NLI judgments (\textit{entail}, \textit{neutral}, \textit{contradict}), enabling automatic evaluation at scale with LLM judges.
    \item We conduct extensive experiments across 13 LLMs and identify key limitations in current models, including evidence of ``\textit{overthinking}'' and a negative association between reasoning length and explanation success.
    \item We show that our task and metrics can be adapted for model training, finding performance gains on two narrative reasoning benchmarks (\textit{True Detective} and \textit{MuSR}).\footnote{We plan to publicly release our code.}
    
\end{enumerate}

\section{Related Work}\label{sec:related_work}

\textbf{LLMs and conflicting information}. While the challenge of handling conflicts between information sources in LLMs has gained significant attention \cite{chen-etal-2022-rich, xie2024adaptive, xu-etal-2024-knowledge-conflicts}, existing work typically frames this as an arbitration task: a process of deciding which facts or conclusions are correct and which should be discarded \cite{liu-roth-2025-conflicts, huo-etal-2025-micro}.\footnote{This contrasts with research on subjective views and opinions, where studies increasingly encourage pluralistic models capable of accommodating multiple contrasting perspectives (see e.g. \citealp{feng-etal-2024-modular}). While we recognize the importance of pluralism in these subjective contexts, such conflicts fall outside the scope of our work.} For example, in BoardgameQA \cite{kazemi2023boardgameqa}, models arbitrate contradictions according to pre-defined heuristics, e.g. preferring conclusions derived from one rule over those derived from another. Our work investigates an alternative to this arbitration paradigm by evaluating how effectively LLMs can instead \textit{reconcile} apparent contradictions.

\textbf{Abductive reasoning in natural language}. $\alpha$-NLI \citet{Bhagavatula2020Abductive} assesses language models' abilities to identify and generate plausible explanations for hypothetical everyday scenarios (and similarly \citet{cao-etal-2026-semeval} for real-world scenarios) but overlooks the role of explanations in reconciling contradictory observations. The most closely related work is \textsc{uncommonsense} \cite{zhao-etal-2024-uncommonsense}, which evaluates model explanations for \textit{unlikely} sequences of commonsense events. While their evaluation methodology relies on pairwise human preferences conflating a range of criteria, we introduce a scalable framework for automatic evaluation grounded in NLI judgments that targets specific qualities necessary for a successful reconciliatory explanation. 

We discuss further related works in Appendix \ref{app:additional_related_work}.

\section{Task formulation}\label{sec:task_formulation}

We follow a conventional definition of ``\textit{contradiction}'' in existing NLP work as occurring ``\textit{when two sentences are extremely unlikely to be true simultaneously}'' \cite{de-marneffe-etal-2008-finding}. We start with the typical setup of an NLI task: given a premise \textit{p} and a hypothesis \textit{h}, a judge model $M_{nli}$ predicts the relationship between \textit{p} and \textit{h}: whether \textit{h} is entailed by \textit{p}, neutral with respect to \textit{p}, or in contradiction with \textit{p}.\footnote{We follow the characterization of ``\textit{entailment}'' in \citet{dagan_textual_entailment} as a human reading \textit{p} would typically infer that \textit{h} is most likely true.} Formally ($\rightarrow$ denoting a mapping):

\begin{quote}
    $M_{nli}(p, h) \rightarrow l$, \; $l \in$ \{\textit{ent.}, \textit{neu.}, \textit{con.}\}
\end{quote}

Given \textit{p} and \textit{h} such that $M_{nli}(p,h) = con.$, the task we formulate is for a model $M_{expl}$ to generate a natural language explanation \textit{e} that is both \textbf{effective} and \textbf{coherent} in reconciling the contradiction with respect to the judge model $M_{nli}$. Specifically, given \textit{p} and \textit{e} combined, we consider \textit{e} \textit{effective} if $M_{nli}$ predicts that \textit{h} is entailed by this combined context; and \textit{coherent} if $M_{nli}$ predicts that \textit{p} and \textit{e} do not contradict each other (per the same definition of ``\textit{contradiction}'' as above).\footnote{As described in Section \ref{subsec:prompting_setup}, we combine \textit{p} with \textit{e} by concatenating them in this order in the context.}  Formally:

\begin{quote}
    $M_{expl}(p, h) \rightarrow e$ such that
    \begin{enumerate}
        \item $M_{nli}(p;e,h) = ent.$ (\textit{effective})
        \item $M_{nli}(p,e) \neq con.$ (\textit{coherent})
    \end{enumerate} 
\end{quote}

Our effectiveness criterion is based on the principle, according to the Inductive-Statistical account of explanations from philosophy of science \cite{Hempel1965-HEMAOS}, that the explanans (statements or claims used in explaining) should make the explanandum (observation being explained) \textit{highly likely}. We therefore implement this principle by requiring that the premise (p) combined with the explanation (e) should \textit{textually entail} the hypothesis (h), i.e. “\textit{a human reading [p and e] would typically infer that [h] is }most likely \textit{true}” \cite{dagan_textual_entailment}. 

Similarly, our coherence criterion is based on the intuition and principle, recognized in \citet{Hempel1965-HEMAOS} and later, e.g. in \citet{FALAPPA20021}, that the explanation itself should also be consistent with (i.e. not contradict) any other statements or observations used in the explaining.

The key strength and novelty of our formulation is that we express the task criteria (\textit{effectiveness} and \textit{coherence}) in terms of standard NLI three-way judgments (\textit{entail}, \textit{neutral}, \textit{contradict}). This facilitates automatic evaluation in two key ways:

First, existing work has found LLM judges to be highly sensitive to complex and novel prompt instructions and scoring rubrics (see e.g. \citealp{thakur-etal-2025-judging}). Accordingly, by prompting models with instructions typically used in existing NLI studies (e.g. in \citet{stacey-etal-2026-improving}), our evaluation approach mitigates the risk of custom-engineered instructions and rubrics pushing the judge model out of distribution.

Second, our formulation provides a principled justification for expecting that stronger NLI performance directly translates into more reliable judgments, meaning that future replications of our study can benefit from continuing advances in NLI to improve judge reliability.

\section{Methodology}\label{sec:experiment}

For each premise–hypothesis pair judged contradictory in an NLI dataset, we prompt a set of LLMs to generate an explanation intended to reconcile the apparent contradiction. The generated explanations are then evaluated by an NLI judge model that has initially labeled these instances as ``\textit{contradiction}''. The key idea is that if this verdict is flipped after the judge model is given the generated explanation, the explanation has successfully reconciled the contradiction. We compute metrics for explanation \textit{effectiveness} (whether the premise combined with the explanation entails the hypothesis), \textit{coherence} (whether the explanation remains consistent with the premise), and overall \textit{success} (i.e. whether the explanation is both effective \textit{and} coherent).

\subsection{REG Dataset Selection}\label{subsec:dataset}

We construct a REG dataset from existing NLI datasets. In this work, we use the subset of MultiNLI \cite{williams-etal-2018-broad} re-annotated as ChaosNLI \cite{nie-etal-2020-learn}, though the approach can in principle be extended to any NLI dataset with contradiction instances. This subset comprises 1,599 instances, each annotated by 100 human annotators (we refer to this subset as \textbf{ChaosNLI-MNLI}). MultiNLI itself is created by first collecting naturally occurring English sentences in diverse genres of spoken and written language (e.g., fiction, letters, conversation transcripts) as premises, and then crowd-sourcing human-written hypotheses. In ChaosNLI-MNLI, each instance consists of a premise-hypothesis pair and a label distribution, denoted as $(w_{ent.}, w_{neu.}, w_{con.})$, representing the proportion of 100 annotators who assigned each respective label. 

From this subset, we use only those instances whose most frequent label is ``\textit{contradiction}'', i.e. $\text{argmax}_{i \in \{ent., neu., con.\}} w_i = w_{con.}$, consisting of 275 instances (we refer to this further subset as \textbf{ChaosNLI-MNLI-C}). 

\begin{figure}
    \centering
    \includegraphics[width=1.0\linewidth]{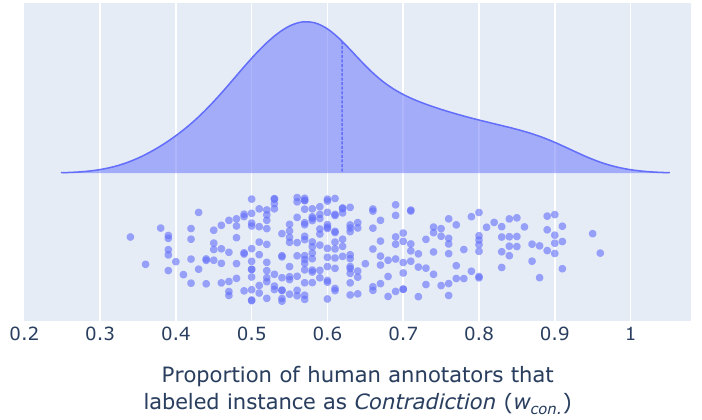}
    \caption{``\textit{Contradiction}'' label weight ($w_{con.}$) of each instance (i.e. proportion of annotators that labeled it as ``\textit{Contradiction}'') in the ChaosNLI-MNLI-C subset (275 instances in total).}
    \label{fig:c_weight_distribution}
\end{figure}

A key rationale in selecting ChaosNLI-MNLI-C for our experiment is that these contradictions are demonstrated to be contentious: as seen in Figure \ref{fig:c_weight_distribution}, none of these instances was unanimously labeled by human annotators as ``\textit{Contradiction}'' (i.e., $w_{con.}=1$), while the mean $w_{con.}$ is 0.62. 

For example, one premise ``\textit{The National Football League semifinals are set.}'' and its corresponding hypothesis ``\textit{They were unable to disclose when the dates would be set.}'' has $w_{con.}=0.7$, meaning only 70 annotators consider this a ``\textit{contradiction}''. A possible explanation to reconcile this apparent contradiction is that while the organizers may have decided on the dates for the semifinals, they might be bound by confidentiality agreements not to announce their decision before giving their sponsors sufficient notice in advance. We discuss more such examples in Appendix \ref{app:more_contradiction_examples}.

\subsection{Explanation Generation}

\paragraph{Prompting.} For each instance in ChaosNLI-MNLI-C, we present the premise \textit{p} and hypothesis \textit{h} to an LLM, and instruct the model to ``\textit{generate a plausible explanation that resolves the apparent contradiction between the premise and the hypothesis}'', stating the requirements for both effectiveness and coherence as set out in Section \ref{sec:task_formulation}. See Appendix \ref{app_sec:prompt_generate_explanations} for the full prompt used. 

\paragraph{Models}\label{subsec:models} We evaluate the explanation generation capability of 13 models, including \textbf{1-5}: Qwen3.5-[0.8B,2B,4B,9B,27B] \cite{qwen3-5}; \textbf{6-10}: Gemma-4-[E2B,E4B,12B,26B-A4B,31B]-it \cite{gemmateam2026gemma4technicalreport}; and \textbf{11-13}: gpt-5.6-[luna,terra,sol] \cite{singh2025openaigpt5card}.

All our selected models support both non-thinking (directly outputting answer) and thinking (first generating a thought trace before outputting answer) modes. We evaluate each model in both modes and report the results accordingly under ``\textit{Non-thinking}'' and ``\textit{Thinking}''.
Further model details are in Appendix \ref{app:model_details}.

\subsection{Reconciliatory Explanation Evaluation}\label{subsec:metrics}

\paragraph{NLI judge model.} To select a judge model, we evaluate the same 13 models as above on their accuracy in predicting the most frequent label for a filtered subset of instances in ChaosNLI-MNLI, where an instance's most frequent label has been assigned by at least a threshold proportion \textit{t} (we set $t=0.8$) out of all annotators.\footnote{Our study uses a single NLI judge since, empirically, we found that the best NLI model Gemma-4-31B-it already exhibits strong performance. Nonetheless, as with LLM-as-a-judge approaches in general, we can also in principle utilize a panel of multiple judges \cite{verga2024replacingjudgesjuriesevaluating}.} 

We base judge selection on ChaosNLI-MNLI for two key advantages. First, compared to other NLI datasets, it offers a rigorous representation of human judgment with 100 annotators per instance. Second, it ensures that the selected judge is competent in assessing NLI relationships involving sentences contained within the same dataset from which we have selected contradictory premise-hypothesis pairs to be reconciled. 

See Appendix \ref{app_sec:prompt_nli} for the full NLI prompt used. Based on the results in Table \ref{tab:nli_judge_accuracy}, we select Gemma-4-31B-it as the NLI judge for our experiment.\footnote{Appendix \ref{sec:additional_judge_selection_results} reports results achieved by each candidate model when other threshold values \textit{t} are used, including $t=0.0$ (using the full dataset unfiltered).} 

To validate the robustness of our chosen judge, we perform additional control experiments by instructing the model to assess explanations that are adversarially crafted to appear relevant but are actually ineffective. Testing both manually written and LLM-generated adversarial explanations, we found that our judge rejects almost all of these explanations as unsuccessful (see Appendix \ref{app:control_experiment_with_adversarial_explanations}). 

\paragraph{Prompting.}\label{subsec:prompting_setup} As defined in Section \ref{sec:task_formulation}, the success of each generated explanation \textit{e} is automatically evaluated by prompting the judge model to predict the NLI label for each of two relationships: (i) between $p;e$ (\textit{p} concatenated with \textit{e})\footnote{We concatenate with the premise first and explanation second primarily for computational efficiency. This is because we expect judge models to evaluate multiple candidate explanations against the same premise, and this order allows the model to pre-compute and reuse the KV cache for the premise tokens, an advantage that scales with longer premises that we expect models to handle in downstream applications.} and \textit{h}; and (ii) between \textit{p} and \textit{e}. In both cases, the same NLI instruction prompt is used as in the judge selection process set out above (and in full in Appendix \ref{app_sec:prompt_nli}). 

\begin{table}[h]
\centering
\small 
\renewcommand{\arraystretch}{1.03}
\begin{tabular}{p{1.3cm} l c}
\midrule
& \textbf{\shortstack{Candidate\\Judge Model}} & \textbf{ } \\
\midrule
\multirow{2}{=}{\textit{\textcolor{gray}{Qwen3.5}}} 
& Qwen3.5-0.8B & 59.17 \\
& Qwen3.5-2B & 62.28 \\
& Qwen3.5-4B & 88.58 \\
& Qwen3.5-9B & 91.35 \\
& Qwen3.5-27B & 93.08 \\
\midrule
\multirow{3}{=}{\textit{\textcolor{gray}{Gemma-4}}} 
& Gemma-4-E2B-it & 74.74 \\
& Gemma-4-E4B-it & 80.97 \\
& Gemma-4-12B-it & 93.08 \\
& Gemma-4-26B-A4B-it & 89.27 \\
& Gemma-4-31B-it & \fcolorbox{blue}{white}{\textbf{94.46}} \\
\midrule
\midrule
\multirow{3}{=}{\textit{\textcolor{gray}{OpenAI\\(proprietary)}}} 
& gpt-5.6-luna & 91.70 \\
& gpt-5.6-terra & 93.08 \\
& gpt-5.6-sol & 92.39 \\[1.0ex]
\midrule
\midrule
\end{tabular}
\caption{Accuracy of candidate judge models (all tested in non-thinking mode) in predicting the most frequently assigned label of instances in ChaosNLI-MNLI with at least 80\% annotator agreement. Score of our chosen judge, Gemma-4-31B-it, is enclosed in \fcolorbox{blue}{white}{\textbf{blue}}.}
\label{tab:nli_judge_accuracy}
\end{table}

\paragraph{Evaluation metrics}
We evaluate generated explanations using a single NLI judge model, $J$. We restrict evaluation to the subset of instances, $N$, where $J$ initially predicts the premise $p_n$ and hypothesis $h_n$ as a contradiction (such that: $\forall n \in N. J(p_n, h_n) = \text{con.}$). With respect to our judge model Gemma-4-31B-it, this subset \textit{N} consists of 195 premise-hypothesis pairs.\footnote{Intuitively, a judge can only validly assess the success of a reconciliatory explanation if it did consider the hypothesis to contradict the premise in the first place.} 

Given an explanation model $M$, we use $N^{M}$ to denote the set of tuples ($p_{n}$, $h_{n}$, $e_{n}^{M}$), where $e_{n}^{M}$ denotes the explanation generated by $M$ for reconciling $p_{n}$ and $h_{n}$. We compute the \textit{effectiveness} (Equation \ref{eq:effectiveness}), \textit{coherence} (Equation \ref{eq:coherence}) and \textit{success} (Equation \ref{eq:success}) scores across all instances in $N_{M}$. As described in Section \ref{sec:task_formulation}, we consider an explanation \textit{e} (i) \textit{effective} if, when the premise \textit{p} is read in conjunction with \textit{e}, this combined context entails the hypothesis; (ii) \textit{coherent} if \textit{e} does not contradict \textit{p}; and (iii) \textit{successful} if \textit{e} is both effective and coherent. Formally:
\vspace{-2pt}
\begin{equation}\label{eq:effectiveness}\small
    \frac{1}{|N^{M}|}\sum_{(p_{n}, h_{n}, e_{n}^{M})}^{N^{M}}\llbracket J(p_{n}; e_{n}^{M}, h_{n}) = ent.\rrbracket
\end{equation}
\vspace{-2pt}
\begin{equation}\label{eq:coherence}\small
    \frac{1}{|N^{M}|} \sum_{(p_{n}, h_{n}, e_{n}^{M})}^{N^{M}} \llbracket 
    \begin{aligned}[t]
    & (J(p_{n}, e_{n}^{M}) \neq con.) \rrbracket
    \end{aligned}
\end{equation}
\vspace{-2pt}
\begin{equation}\label{eq:success}\small
    \frac{1}{|N^{M}|} \sum_{(p_{n}, h_{n}, e_{n}^{M})}^{N_{J}^{M}} \llbracket 
    \begin{aligned}[t]
    & (J(p_{n}; e_{n}^{M}, h_{n}) = ent.) \\
    & \wedge (J(p_{n}, e_{n}^{M}) \neq con.) \rrbracket
    \end{aligned}
\end{equation}
where square brackets denote an indicator function (1 if true, 0 otherwise).

\subsection{Implementation}\label{subsec:implementation}

For all our experiments in this study (including the NLI judge selection process), we use the default sampling parameters recommended by \citet{qwen3-5} when prompting models in non-thinking and thinking modes respectively.\footnote{See \url{https://huggingface.co/Qwen/Qwen3.5-27B} sampling parameters corresponding to ``[non-thinking/thinking] mode for general tasks''.} We set reasoning effort at ``\textit{medium}'' when prompting gpt-5.6-[luna,terra,sol] in thinking mode. Our NLI judge is prompted in non-thinking mode throughout our experiments. See Appendix \ref{app:sampling_params} for details.

We use the vllm implementation of all open-weight models as of v0.26.0, \cite{kwon2023efficientmemorymanagementlarge_vllm}. All experiments are run on NVIDIA H100-NVL GPUs. OpenAI models are accessed via API.

\section{Results}\label{sec:main_results}

\begin{table*}[t]
\centering
\footnotesize 
\renewcommand{\arraystretch}{1}
\setlength{\tabcolsep}{2pt}
\begin{tabular}{p{1.3cm} l c c c c c c}
\hline
& \multirow{2}{*}{\textbf{\shortstack{Source Model\\of Generated\\Explanations}}} & \multicolumn{2}{c}{\textbf{Coherence}} & \multicolumn{2}{c}{\textbf{Effectiveness}} & \multicolumn{2}{c}{\textbf{Explanation Success}} \\
\cmidrule(lr){3-4} \cmidrule(lr){5-6} \cmidrule(lr){7-8}
& & \textbf{\shortstack{Non-\\Thinking}} & \textbf{Thinking} & \textbf{\shortstack{Non-\\Thinking}} & \textbf{Thinking} & \textbf{\shortstack{Non-\\Thinking}} & \textbf{Thinking} \\
\hline
\multirow{5}{=}{\textit{\textcolor{gray}{Qwen 3.5}}} 
& Qwen3.5-0.8B & 72.31 & 72.82 & 26.67 & 22.05 & 11.28 & 16.41 \\
& Qwen3.5-2B   & 66.15 & 66.15 & 38.46 & 57.95 & 18.46 & 33.33 \\
& Qwen3.5-4B   & 76.41 & 74.36 & 17.95 & 57.44 & 13.33 & 40.51 \\
& Qwen3.5-9B   & 86.67 & 81.54 & 30.26 & 64.10 & 24.62 & 51.28 \\
& Qwen3.5-27B  & 75.90 & 80.51 & 61.54 & 70.77 & 44.10 & 54.36 \\
\hline
\multirow{5}{=}{\textit{\textcolor{gray}{Gemma 4}}} 
& Gemma-4-E2B-it     & 87.18 & 82.05 & 26.67 & 30.77 & 20.00 & 23.08 \\
& Gemma-4-E4B-it     & \textbf{90.77} & 85.13 & 18.46 & 38.46 & 13.33 & 31.28 \\
& Gemma-4-12B-it     & 82.56 & 80.51 & 43.59 & 57.95 & 30.77 & 46.67 \\
& Gemma-4-26B-A4B-it & 80.51 & 85.13 & 47.69 & 55.38 & 34.36 & 44.10 \\
& \fcolorbox{blue}{white}{Gemma-4-31B-it}     & 81.54 & 81.03 & 57.44 & 62.05 & 45.64 & 47.18 \\
\hline\hline
\multirow{3}{=}{\textit{\textcolor{gray}{OpenAI\\(propri-\\etary)}}} 
& gpt-5.6-luna  & 81.54 & \textbf{89.23} & 49.23 & 70.26 & 35.38 & 62.05 \\[1ex]
& gpt-5.6-terra & 84.10 & 81.03 & 55.90 & 77.44 & 44.10 & 61.03 \\[1ex]
& gpt-5.6-sol   & 85.64 & 80.00 & \textbf{73.33} & \textbf{79.49} & \textbf{62.05} & \textbf{62.56} \\[1ex]
\hline\hline
\end{tabular}
\caption{Coherence, effectiveness, and success rate of each model's generated explanations (in \%), as evaluated by Gemma-4-31B-it as the judge model (enclosed in \fcolorbox{blue}{white}{blue}). Scores in the row corresponding to Gemma-4-31B-it are when the model itself acts as both explanation generator and the judge doing the evaluating. Highest score in each column is in \textbf{bold}. Appendix \ref{app:different_judge_exp} includes an extended version of this table with 95\% bootstrap confidence intervals.}
\label{tab:explanation_metrics_full}
\end{table*}

As shown in Table \ref{tab:explanation_metrics_full}, most models are reasonably capable of generating coherent explanations that do not contradict the premise. Gemma-4-E4B-it in non-thinking mode achieves the highest coherence score of 90.77, while Qwen3.5-2B performs the worst across both thinking and non-thinking modes at 66.15. Surprisingly, 8 out of 13 models perform slightly worse in \textbf{coherence} when prompted in thinking versus non-thinking mode, with the largest drop being 5.64 in Gemma-4-E4B-it. 

By contrast, generating effective explanations is more difficult for the majority of models, with performance ranging from 17.95 (Qwen3.5-4B in non-thinking mode) to 79.49 (gpt-5.6-sol in thinking mode). Since an explanation is considered successful only if it is both coherent and effective, this means that models also exhibit mediocre performance in generating successful explanations. Except for the proprietary models (gpt-5.6-[luna,terra,sol]) in thinking mode, none of our models achieve a success rate of over 55, with the closest being Qwen3.5-27B (54.36) and Gemma-4-31B-it (47.18) respectively. 

To assess the reliability of our LLM judge in assessing explanation success, two authors independently annotated a subset of (N=104) generated explanations. As detailed in Appendix \ref{app:comparing_humans_v_llm_judge}, agreement between the LLM judge and each human annotator is moderate (Cohen's kappa \cite{Cohen1960} $\kappa \in [0.374, 0.392]$) and comparable to human–human agreement ($\kappa = 0.427$). Among the 79 examples on which both human annotators agree, our LLM judge matches the agreed human label with 82.3\% accuracy (MCC = 0.522).

To further validate our findings, we also replicate the experiment using a different model (gpt-5.6-sol) as judge and compare these results in Appendix \ref{app:different_judge_exp}.

\section{Analysis}

As shown in Table \ref{tab:n_token_stats}, when prompted in thinking mode, we find that models of different families differ widely in the number of reasoning tokens they produce in generating an explanation, with the most verbose being models in the Qwen 3.5 family, and the least being those in the gpt-5.6 family (using on average around 60 times fewer tokens than the Qwen 3.5 models). Across all models, we also observe that unsuccessful explanations (as assessed by our LLM judge) on average took more tokens to generate than successful ones. 

A possible explanation is that this pattern of unsuccessful explanations consuming more tokens reflects ``\textit{overthinking}'', whereby longer reasoning does not improve (and can even harm) task performance \cite{chen2025do,wu2026when,fan2025missing}. However, we need to account for the possibility that models may also simply reason for longer where the premise-hypothesis pair is harder to reconcile in the first place. 

To investigate this, we fit a logistic generalized linear model \cite{glm_book} predicting whether an explanation is successful based on output length (including reasoning tokens)\footnote{We apply $log_{2}$ transformation to no. of output tokens.}, the specific explanation model and the contradiction weight ($w_{con.}$) assigned by annotators in \citet{nie-etal-2020-learn}. Intuitively, human annotators may be more likely to judge a premise-hypothesis pair as contradictory when they cannot readily think of an explanation or scenario in which both statements could hold true (as in studies of human reasoning, e.g. \citealp{doi:10.1073/pnas.1012933107}). We therefore use the weight assigned to the ``\textit{contradiction}'' label ($w_{con.}$) as a proxy for the difficulty in reconciling the premise and hypothesis. We restrict our analysis to explanations generated in thinking mode, and cluster standard errors by premise-hypothesis pair. 

First, in line with our intuition set out above, we find that higher contradiction weight is associated with lower explanation success, with a 0.1 increase in $w_{con.}$ corresponding to 12.6\% lower odds of success (OR = 0.87, 95\% CI [0.78, 0.98], \textit{p} = .016). Then, specifically with respect to explanations generated in thinking mode, controlling for $w_{con.}$ and the specific explanation model, we find that doubling output length is associated with 44.3\% lower odds of success (OR = 0.56, 95\% CI [0.45, 0.68], \textit{p}<.001). Thus, we find that \textbf{longer reasoning length is negatively correlated with explanation success}, and that this persists even after accounting for the difficulty in reconciling the specific premise-hypothesis pair.

\begin{table} 
\centering 
\small 
\setlength{\tabcolsep}{2pt} 
\renewcommand{\arraystretch}{1.03} 
\begin{tabular}{p{1.1cm} p{2.7cm} c c} 
\midrule 
 & \multirow{2}{*}{\textbf{\shortstack{Source Model\\of Generated\\Explanations}}} & \multicolumn{2}{c}{\textbf{mean \textit{\#} of output tokens}} \\
\cmidrule(lr){3-4}
 & & \textbf{\shortstack{successful\\explanations}} & \textbf{\shortstack{unsuccessful\\explanations}} \\ 
\midrule 
\multirow{5}{=}{\textit{\textcolor{gray}{Qwen3.5}}} & Qwen3.5-0.8B & 6211.84 & 6470.07 \\ 
& Qwen3.5-2B & 5757.32 & 6173.87 \\ 
& Qwen3.5-4B & 5596.57 & 6153.34 \\ 
& Qwen3.5-9B & 5324.74 & 5409.26 \\ 
& Qwen3.5-27B & 4905.58 & 5542.65 \\ 
\midrule 
\multirow{5}{=}{\textit{\textcolor{gray}{Gemma-4}}} & Gemma-4-E2B-it & 509.71 & 523.18 \\ 
& Gemma-4-E4B-it & 669.92 & 707.34 \\ 
& Gemma-4-12B-it & 2031.88 & 2544.56 \\ 
& \shortstack[l]{Gemma-4-26B-\\A4B-it} & 3065.27 & 3797.07 \\ 
& Gemma-4-31B-it & 1043.45 & 1585.96 \\ 
\midrule \midrule 
\multirow{3}{=}{\textit{\textcolor{gray}{OpenAI\\\shortstack{(proprie-\\tary)}}}} & gpt-5.6-luna & 178.07 & 181.65 \\ 
& gpt-5.6-terra & 116.79 & 160.21 \\ 
& gpt-5.6-sol & 118.30 & 121.90 \\[1.0ex] 
\midrule \midrule 
\end{tabular} 
\caption{Comparison of mean output tokens (including reasoning tokens) generated by each model in thinking mode, for successful and unsuccessful explanations.} 
\label{tab:n_token_stats} 
\vspace{-0.143in}
\end{table}

We discuss results of our qualitative analysis in Appendix \ref{app:qualitative_analysis} (with examples of generated explanations in Tables \ref{tab:qual_analysis_p1} and \ref{tab:qual_analysis_p2}) including failure cases in non-thinking and thinking modes, such as superficial fluency, error propagation and overthinking.

\section{Adapting REG for Model Training}\label{sec:model_training}

Existing work shows that training models on \textit{fundamental} reasoning skills in symbolic and logical domains can lead to generalizations that improve downstream task performance \cite{NEURIPS2024_8678da90,cao-etal-2026-fundamental}. Notably, however, training models on a formal, rule-based abduction task transfers poorly to realistic natural language tasks, as compared to training on formal deduction or induction \cite{cao-etal-2026-fundamental}. We speculate that this is in part due to a mismatch between symbolic premises used in training, and ambiguous, implication-rich natural language contexts over which abductive reasoning is required in more realistic settings.

As such, we conduct a proof-of-concept experiment to test whether our task of reconciliatory explanation generation, which requires abductive reasoning directly in natural language, together with our proposed automatic metrics for assessing such explanations, can be adapted into a viable training regime to improve models' reasoning capabilities.

\subsection{Experiment Setup}

Using the same dataset ChaosNLI-MNLI-C, we train Qwen3.5-2B on our task of reconciliatory explanation generation with Decoupled Clip and Dynamic Sampling Policy Optimization (DAPO) \cite{yu2025dapo}, a more recently proposed improvement of Group Relative Policy Optimization (GRPO) \cite{grpo_paper}. In particular, its ``\textit{clip-higher}'' mechanism is designed to mitigate entropy collapse and promote exploration by increasing the upper clipping threshold of the importance-sampling ratio, thereby allowing larger increases in the probability of low-probability tokens with positive advantages.\footnote{Specifically, DAPO decouples the lower and upper clipping thresholds of the importance sampling ratio, using $1-\epsilon_{\mathrm{low}}$ and $1+\epsilon_{\mathrm{high}}$, respectively, with $\epsilon_{\mathrm{high}}>\epsilon_{\mathrm{low}}$. This relaxes the upper clipping constraint on policy updates that increase token probabilities. See \citet{yu2025dapo} for further details and the full algorithm.} We choose to use on-policy reinforcement learning over supervised fine-tuning in order to reduce distribution mismatch between training and inference, and to reflect the one-to-many nature of our task, in that each premise-hypothesis pair can be successfully reconciled with multiple distinct explanations.

During the rollout phase, we prompt the model in non-thinking mode to generate a reconciliatory explanation \textit{e} for a contradictory premise-hypothesis pair in ChaosNLI-MNLI-C, and sample 8 outputs from the model. Per Table \ref{tab:rl_metric}, each output is assigned a positive reward if \textit{e} is judged (by Gemma-4-31B-it) as both effective \textit{and} coherent, a negative reward if either \textit{e} contradicts the premise, or the premise and \textit{e} combined contradicts the hypothesis; and 0 otherwise. In particular, because models in our main evaluation appear more capable of generating coherent than effective explanations, we design the reward function to reward coherence only when the explanation is also effective, to mitigate the risk that a model over-optimizes on coherence, an easier objective, at the expense of effectiveness. To provide models with more fine-grained learning signals, our rubric also assigns a higher positive reward to coherent explanations that are textually entailed by the premise (1), compared to coherent explanations that merely do not contradict it (0.5).

We reuse the same prompts in our earlier evaluation (given in full in Appendix \ref{app:full_prompts_used}) to elicit explanations and NLI judgments, respectively. We train models using the TRL library \cite{von_Werra_TRL_Transformers_Reinforcement_2020} with the default hyperparameters (see Appendix \ref{app:training_hyperparams}), and only varying the upper clip of the importance sampling ratio ($\epsilon_{high}\in[0.2, 0.28, 0.36]$), given our computational budget. 
\begin{table}[]
\footnotesize
\centering
\begin{tabular}{lllll}
\hline
\multicolumn{2}{l}{\multirow{2}{*}{}} & \multicolumn{3}{l}{\shortstack{Effectiveness\\$M_{NLI}(p;e,h)$}} \\ \cline{3-5} 
\multicolumn{2}{l}{} & \textit{con.} & \textit{neu.} & \textit{ent.} \\ \hline
\multirow{3}{*}{\shortstack{Coherence\\$M_{NLI}(p,e)$}} & \textit{con.} & -1 & -1 & -1 \\ \cline{2-5} 
 & \textit{neu.} & -1 & 0 & 0.5 \\ \cline{2-5} 
 & \textit{ent.} & -1 & 0 & 1 \\ \hline
\end{tabular}
\caption{Reward matrix for RL, based on NLI judge's evaluation of explanation coherence $M_{NLI}(p, e)$ and effectiveness $M_{NLI}(p; e,h)$. Unlike task metrics used in evaluation, we introduce negative (-1) and partial (0.5) rewards based on specific NLI label combinations.}
\label{tab:rl_metric}
\vspace{-0.05in}
\end{table}

We evaluate the trained models on True Detective \cite{del-fishel-2023-true}, a benchmark of 191 detective puzzles designed to assess abductive reasoning; and MuSR \cite{sprague2024musr}, a dataset for assessing models' general multistep soft reasoning over long narratives.\footnote{We use the ``\textit{murder mystery}'' subset of MuSR consisting of 250 instances.} 

For each of the three hyperparameter configurations (differing only in $\epsilon_{high}$), we perform three training runs using different random seeds and report the mean evaluation score across runs, with each trained model evaluated in non-thinking mode. We compare against baseline scores of Qwen3.5-2B prompted in non-thinking and thinking modes.

\subsection{Results}

\begin{table}[t]
\centering
\footnotesize
\setlength{\tabcolsep}{3.5pt}
\renewcommand{\arraystretch}{1.1}
\begin{tabular}{@{}llcc@{}}
\hline
\textbf{Model} &
\textbf{Mode} &
\shortstack{\textbf{True}\\\textbf{Detective}} &
\textbf{MuSR} \\
\hline
\multirow{2}{*}{Baseline}
& Non-thinking & 22.51 & 38.40 \\
& Thinking     & \textbf{25.65} & \textbf{57.20} \\
\hline
DAPO-trained & & & \\
\hspace{1em}$\epsilon_{\mathrm{high}}=0.20$
& \multirow{3}{*}{Non-thinking} & 23.21 & \underline{43.47} \\
\hspace{1em}$\epsilon_{\mathrm{high}}=0.28$
& & \underline{24.61} & 40.53 \\
\hspace{1em}$\epsilon_{\mathrm{high}}=0.36$
& & 23.73 & 43.20 \\
\hline
\end{tabular}
\caption{Accuracy (\%) on True Detective and MuSR of the baseline Qwen3.5-2B (in non-thinking and thinking modes) and DAPO-trained models (only in non-thinking mode) with different upper clipping thresholds $e_{high}$. Accuracy of DAPO-trained models are mean values across three training runs with same hyperparameters and different random seeds. Highest score in each column is shown in \textbf{bold} and second highest \underline{underlined}.}
\label{tab:training_results}
\end{table}

As shown in Table \ref{tab:training_results}, all trained Qwen 3.5 models in non-thinking mode outperform the non-thinking baseline across both benchmarks, with the most substantial gain being +5.07 in percentage accuracy on MuSR. Notably, optimal upper clipping $\epsilon_{\mathrm{high}}$ differ across tasks, with $\epsilon_{\mathrm{high}}=0.28$ performing best on True Detective, while $\epsilon_{\mathrm{high}}=0.20$ best on MuSR. Although the thinking-mode baseline still achieves the highest scores overall, our pilot results demonstrate that this performance gap between thinking and non-thinking mode in abductive reasoning can in principle be narrowed through training models on our introduced task and automatic metrics. We leave more extensive hyperparameter tuning and analysis, as well as investigating different model sizes and families, for future work.

\section{Conclusion}

We introduce the task of reconciling contradictory observations through natural language explanations, an underexplored but key aspect of intelligence in tasks ranging from conversations to scientific discovery. Our proposed methodology effectively leverages existing NLI datasets and introduces metrics for scalable automatic evaluation with LLMs as judges. Our study leads to new insights into the reasoning abilities of current LLMs, particularly that generating reconciliatory explanations remains challenging and that longer ``thinking'' on this task is associated with worse performance. Further, our proof-of-concept experiment shows how our introduced task and metrics can be adapted for reinforcement-training for model improvement. Altogether, our work highlights a challenge in current LLMs' reasoning capabilities and a promising direction for this to be addressed in future work.

\section*{Limitations}

Whilst our proposed task is language-agnostic in principle, our current experiments only evaluate models' reconciliation capabilities using premise-hypothesis pairs in English. Evaluating models across diverse languages is bottlenecked by the availability of high-quality multilingual NLI datasets, though recent work indicates that this challenge is being addressed by the community (see e.g. \citealp{vrabcova-etal-2025-towards}).

Additionally, our study restricts its focus to generating explanations for relatively short premise-hypothesis pairs, typically spanning no more than one or two sentences. This has enabled our evaluation to specifically target models' reconciliatory explanation generation capability, isolated from other capabilities such as factual knowledge retrieval and long-context reasoning. Nonetheless, because real-world contradictions could emerge across longer contexts, future work could also expand on this framework to evaluate models on document-level or multi-hop contradictions.

Further, while we had manually inspected explanations generated by models on our task, our study intentionally avoids introducing a formal taxonomy to categorize these explanations. As well-recognized in ongoing debates across psychology, philosophy, and linguistics \cite{Keil2006-ol, Woodward2003-WOOMTH, Lombrozo2006}, establishing a universal taxonomy for explanations is notoriously challenging due to the inherent nature of explanations being highly context-dependent and exhibiting wide varieties. This is especially the case given the diverse sources, genres and subject matters of the main dataset ChaosNLI-MNLI in our study. 

Whilst our proof-of-concept experiment demonstrates the \textit{feasibility} of adapting our introduced task and metrics for reinforcement-learning training, a more systematic evaluation, including hyperparameter tuning, analysis and comparisons against contemporaneous methods, remains for future work to fully characterize the efficiency, strengths and potential trade-offs of our approach. 

Finally, our task criteria focus purely on the relational properties of the generated explanations with respect to the premise and hypothesis to be explained, to the exclusion of other potentially desirable explanation qualities. 
This approach is informed by the typical characterization of abductive reasoning as a two-stage process of first generating candidate explanations and then selecting the best explanation(s) \cite{Lipton2003-LIPITT-3}, within which our task is a targeted assessment of LLMs’ ability to perform the first stage of this process. In downstream real-world applications, we would expect these model-generated and hypothesized explanations to then be validated through additional retrievals and reasoning, and selected based on additional criteria such as groundedness and novelty. However, these additional criteria can often compete against one another (e.g. conjecturing more novel explanations as scientific hypotheses for testing may require looser grounding on currently available evidence), and need to be informed by application-specific considerations. Evaluating explanations for these additional qualities therefore remains an important avenue for future work. 

\section*{Acknowledgment}

This work was supported by the UKRI AI Centre for Doctoral Training in Speech and Language Technologies (SLT) and their Applications funded by UK Research and Innovation [grant number EP/S023062/1]. For the purpose of open access, the author has applied a Creative Commons Attribution (CC BY) licence to any Author Accepted Manuscript version arising. We acknowledge IT Services at The University of Sheffield for the provision of services for High Performance Computing.

\bibliography{custom}

\appendix

\section{Additional Related Work}\label{app:additional_related_work}

\textbf{Defeasible NLI.} $\delta$-NLI \cite{rudinger-etal-2020-thinking} assesses whether language models can generate contexts that increase (or decrease) the likelihood of a \textit{neutral} hypothesis given a premise scenario. Our work expands this direction by considering contradictory pairs and reconciliatory contexts, and proposing automatic evaluation metrics based not on n-gram overlaps but on NLI judgments.

\textbf{Human label variations in NLI}. Existing work demonstrates that ground truth labels in certain NLI datasets are highly contentious with low annotator agreement (\citealp{nie-etal-2020-learn}; \citealp{10.1613/jair.1.12752}). As \citet{jiang-marneffe-2022-investigating} reveals, these disagreements can stem from a range of valid causes such as differing interpretations of implicature, inherent linguistic ambiguity and ``\textit{under-specification}''. Our work leverages these genuine disagreements as a feature and not a bug, by introducing the task of uncovering a latent context or interpretation that renders the apparent contradiction compatible.  

\textbf{NLI label-flipping}. To assess and improve a model's robustness in NLI tasks, existing work makes minimal counterfactual edits to premises or hypotheses capable of inducing changes in its predicted labels (\citealp{zheng-zhu-2023-natlogattack}; \citealp{wang-etal-2025-truth}, etc.). Our work differs in that we treat the original premises and hypotheses as given observations about the world and, instead, require models to generate \textit{additional} context to induce label change. This reflects a fundamental difference in our evaluation target, which is the reconciliatory capability of explanation models, not the discriminatory robustness of NLI models.

\section{Additional Examples from ChaosNLI-MNLI-C}\label{app:more_contradiction_examples}

\textbf{Example 1 (uid: 12870c)}

Premise (P): \textit{so he donates a lot not everything but a lot of the material then what he doesn't donate we just go out and buy}

Hypothesis (H): \textit{He donates all of the material.}

(89 out of 100 annotators label as contradiction)

If ``all of the material'' is read from the receiver’s perspective (i.e. the donor donated all that the receiver needed), then H appears to contradict P; however, if the phrase is read from the donor’s perspective (i.e. the donor donated all that he had, even though it didn’t cover all that the receiver needed), H could very well be true and compatible with P.

\textbf{Example 2 (uid: 46198c)}

P: \textit{How effectively DOD manages these funds will determine whether it receives a good return on its investment.}

H: \textit{The DOD is certain to have a bad return on these funds.}

(52 out of 100 annotators label as contradiction)

If the reader is agnostic about the DOD’s effectiveness in managing funds, then H appears to contradict P since the DOD might manage these funds well and get a good return; however, if the reader has any preconceived beliefs about government agencies (reading ``DOD'' as referring to ``Department of Defense'') being frequently ineffective in managing their finances despite their autonomy to do so, then they may not recognize H as contradicting P.

\textbf{Example 3 (uid: 134769n)}

P: \textit{But of course the DSM is informed by social values.}

H: \textit{The DSM is mostly concerned with medical inputs rather than social values.}

(61 out of 100 annotators label as contradiction)

A skeptical reader could interpret P as the DSM only paying lip service to considering social values. In other words, while it is technically true that the DSM does consider social values to some degree (meaning P is true), this consideration is far outweighed by medical inputs (making H also true).

\section{Full Prompts}\label{app:full_prompts_used}

\subsection{Instruction Prompt for Generating Explanations}\label{app_sec:prompt_generate_explanations}

\begin{promptbox}

Given a premise and a hypothesis, your task is to generate a plausible explanation that reconciles the apparent contradiction between the premise and the hypothesis. Specifically, when the explanation is combined with the premise, the hypothesis should follow from this combined context.

The explanation should be succinct and no more than a single sentence long. It should not trivially repeat the premise or the hypothesis and must not itself contradict the premise. The explanation should also read naturally in the context as a continuation of the premise, without explicitly using the term 'premise' or 'hypothesis' to refer to the premise or hypothesis.

Your final output should be a single sentence without any preamble.

Premise: \{\{ premise \}\}

Hypothesis: \{\{ hypothesis \}\}

\end{promptbox}

\subsection{Instruction Prompt for NLI Judgments}\label{app_sec:prompt_nli}

\begin{promptbox}
Given a premise and a hypothesis, your task is to judge the relationship between the two texts.
Specifically, you will need to assign one of three labels to the relationship:

- Entailment: The hypothesis is most likely true given the premise.

- Contradiction: The hypothesis is most likely false given the premise.

- Neutral: The hypothesis is not entailed by the premise but does not contradict it either.

Your final output should be only one word, namely the label.

Premise: \{\{ premise \}\}

Hypothesis: \{\{ hypothesis \}\}
\end{promptbox}

As described in Section \ref{subsec:metrics}, automatic evaluation is carried out by an LLM judge predicting the following relationships: (i) between $p;e$ (\textit{p} concatenated with \textit{e}) and \textit{h}, and (ii) between \textit{p} and \textit{e}.

In case of (i), $p;e$ would fill the ``\textit{premise}'' placeholder in the above template, and \textit{h} would fill the ``\textit{hypothesis}'' placeholder.

In case of (ii), \textit{p} would fill the ``\textit{premise}'' placeholder in the above template, and \textit{e} would fill the ``\textit{hypothesis}'' placeholder.

\section{Model Details}\label{app:model_details}
Details of the 13 models used in our experiments are listed in Table \ref{tab:models}.
\begin{table}[htb]
\centering
\small
\renewcommand{\arraystretch}{1.1}
\begin{tabular}{ll}
\toprule
\textbf{Full model name} & \textbf{Parameter count} \\
\midrule
\multicolumn{2}{l}{\textit{\textcolor{gray}{Qwen3.5 \cite{qwen3-5}}}} \\[2pt]
Qwen3.5-0.8B & $8.0 \times 10^8$ \\
Qwen3.5-2B & $2.0 \times 10^9$ \\
Qwen3.5-4B & $4.0 \times 10^9$ \\
Qwen3.5-9B & $9.0 \times 10^9$ \\
Qwen3.5-27B & $2.7 \times 10^{10}$ \\
\midrule
\multicolumn{2}{l}{\textit{\textcolor{gray}{Gemma-4 \cite{gemmateam2026gemma4technicalreport}}}} \\[2pt]
Gemma-4-E2B-it & $5.1 \times 10^9$ \\
Gemma-4-E4B-it & $8.0 \times 10^9$ \\
Gemma-4-12B-it & $1.2 \times 10^{10}$ \\
Gemma-4-26B-A4B-it & $2.5 \times 10^{10}$ \\
Gemma-4-31B-it & $3.1 \times 10^{10}$ \\
\midrule
\multicolumn{2}{l}{\textit{\textcolor{gray}{OpenAI GPT-5.6 \cite{singh2025openaigpt5card}}}} \\[2pt]
gpt-5.6-luna & Undisclosed \\
gpt-5.6-terra & Undisclosed \\
gpt-5.6-sol & Undisclosed \\
\bottomrule
\end{tabular}
\caption{Details of the 13 models used in our experiments. All models can be prompted in both thinking and non-thinking modes.}
\label{tab:models}
\end{table}

\section{Additional Results from Judge Selection}\label{sec:additional_judge_selection_results}

We conduct further analysis with respect to our judge selection process by varying the annotator-agreement threshold \textit{t} which we use to filter ChaosNLI-MNLI instances on which to assess candidate judge models. That is, we only assess models on instances where the most frequent NLI label has been been assigned by at least a proportion \textit{t} out of all annotators.

\begin{table}
\centering
\small

\newcommand{\wrap}[1]{\begin{tabular}[t]{@{}l@{}}#1\end{tabular}}

\begin{tabular}{lcccc}
\toprule
\textbf{\wrap{Candidate\\Judge Model}} & \multicolumn{4}{c}{\textbf{Agreement Threshold ($t$)}} \\
\cmidrule(lr){2-5}
 & \textbf{0.0\textsuperscript{*}} & \textbf{0.7} & \textbf{0.8} & \textbf{0.9} \\
\midrule
Qwen3.5-0.8B                & 43.03 & 52.07 & 59.17 & 70.18 \\
Qwen3.5-2B                  & 48.03 & 57.07 & 62.28 & 66.67 \\
Qwen3.5-4B                  & 69.29 & 85.17 & 88.58 & 96.49 \\
Qwen3.5-9B                  & 69.61 & 85.00 & 91.35 & 100.00 \\
Qwen3.5-27B                 & 72.61 & 90.17 & 93.08 & 100.00 \\
Gemma-4-E2B-it              & 61.35 & 72.41 & 74.74 & 80.70 \\
Gemma-4-E4B-it              & 63.91 & 76.21 & 80.97 & 85.96 \\
Gemma-4-12B-it              & 70.23 & 87.24 & 93.08 & 96.49 \\
\wrap{Gemma-4-\\26B-A4B-it} & 68.92 & 84.31 & 89.27 & 96.49 \\
\textbf{Gemma-4-31B-it}     & \textbf{75.05} & \textbf{91.21} & \textbf{94.46} & \textbf{100.00} \\
gpt-5.6-luna                & 68.17 & 85.86 & 91.70 & 100.00 \\
gpt-5.6-terra               & 71.86 & 88.97 & 93.08 & 98.25 \\
gpt-5.6-sol                 & 71.29 & 88.28 & 92.39 & 98.25 \\
\bottomrule
\end{tabular}
\caption{Accuracy (\%) of candidate NLI judge models on ChaosNLI-MNLI instances, filtered by different annotator agreement thresholds ($t$). All models are prompted in non-thinking mode. The row corresponding to our selected judge, \textbf{Gemma-4-31B-It} is in \textbf{bold}.
\textsuperscript{*}Where $t=0.0$, the full unfiltered dataset is used.}
\label{tab:chaosnli_thresholded_judge_results}
\end{table}

As shown in Table \ref{tab:chaosnli_thresholded_judge_results}, our selected judge Gemma-4-31B-it consistently achieves the highest accuracy across all models in our evaluation. As with other models, the accuracy of Gemma-4-31B-it increases monotonically with the threshold level of annotator agreement used, reaching perfect accuracy when evaluated only on instances with at least 0.9 annotator agreement.

\section{Testing NLI Judge Robustness against Adversarial Explanations}\label{app:control_experiment_with_adversarial_explanations}

We conduct a control experiment to validate that our selected NLI judge model, Gemma-4-31B-it, is robust against adversarial explanations that appear to be relevant, but do not actually reconcile the apparent contradiction between a premise and hypothesis. 

We create these adversarial explanations using two methods. First, we prompt gpt-5.6-terra to generate an explanation for each instance in ChaosNLI-MNLI-C, using this instruction: ``\textit{Given a premise and a hypothesis, your task is to generate an explanation that appears to be relevant to both the premise and hypothesis but does not actually resolve the apparent contradiction between the two. \textbf{It should selectively reuse words and concepts that have appeared in either or both the premise and hypothesis.}}''

In addition, we sample 50 unique premise-hypothesis pairs from ChaosNLI-MNLI-C, and manually write an adversarial explanation for each of these pairs. As per the same instruction we had given to gpt-5.6-terra, we repeat and combine words and concepts in the premise and hypothesis in order to produce deliberately adversarial explanations that are relevant-sounding but ineffective. Table \ref{tab:adversarial_explanations} shows example explanations created with these methods for a premise-hypothesis pair.

\begin{table}[]
\centering
\small
\begin{tabularx}{\linewidth}{l X} 
\toprule
\multicolumn{2}{p{\dimexpr\linewidth-2\tabcolsep}}{%
\textbf{Premise:} \textit{In the depths of the Cold War, many Americans suspected Communists had infiltrated Washington and were about to subvert our democracy.} \newline
\textbf{Hypothesis:} \textit{Communists assisted America's government during the Cold War.} \newline
\footnotesize (57 out of 100 annotators labeled as contradiction)
} \\ 
\midrule \addlinespace[0.5ex]
\textbf{Source} & \textbf{Adversarial Explanation} \\ \midrule
gpt-5.6-terra          & Cold War debates in Washington often blurred the line between government service and fears of political subversion.
 \\ \midrule

\begin{tabular}[t]{@{}l@{}}Manually\\written\end{tabular} & As such, they were eager to assist America's government by reporting any suspected Communists. \\ \bottomrule
\end{tabularx}
\caption{An example premise-hypothesis pair for which adversarial explanations, seemingly relevant but ineffective, are created through (a) prompting gpt-5.6-terra; and (b) manual writing.}
\label{tab:adversarial_explanations}
\end{table}

We then test our judge model on these explanations, calculating the coherence, effectiveness and success rate based on its NLI judgments. 

As shown in Table \ref{tab:judge_robustness_against_adversarial_explanations}, our judge model, Gemma-4-31B-it, mostly rejects these adversarial explanations as unsuccessful, erring only around 2-6\% of the time. Compared to model-generated explanations, the model appears to be slightly more susceptible (+3.95) towards accepting manually written adversarial examples as effective and successful. Nonetheless, taken altogether, We interpret these results as demonstrating that \textit{our judge model, when used according to our proposed criteria and methodology, is sufficiently robust (though imperfect) in its assessment of explanations.}

\begin{table}[]
\centering
\small
\begin{tabularx}{\columnwidth}{@{} X >{\centering\arraybackslash}p{0.8cm} >{\centering\arraybackslash}p{0.8cm} >{\centering\arraybackslash}p{0.8cm} @{}}
\toprule
\textbf{Source of adversarial explanations} & \textbf{C} & \textbf{E} & \textbf{S} \\
\midrule
Model-generated (gpt-5.6-terra)    & 99.49  & 2.05   & 2.05   \\
\addlinespace
Manually written             & 100.00  & 6.00 & 6.00 \\
\bottomrule
\end{tabularx}
\caption{Coherence (\textbf{C}), effectiveness (\textbf{E}) and success rate (\textbf{S}) as predicted by our selected judge, Gemma-4-31B-it, when the model is tasked with evaluating adversarial explanations that are deliberately crafted to appear relevant but ineffective in reconciling the premise and hypothesis.}
\label{tab:judge_robustness_against_adversarial_explanations}
\end{table}

\begin{table*}[t]
\footnotesize
\setlength{\tabcolsep}{4pt}
\begin{tabular}{p{0.16\linewidth} p{0.77\linewidth} c}
\toprule
\textbf{Model} & \textbf{Explanation} & \textbf{E / C / S} \\
\midrule

\multicolumn{3}{p{0.95\linewidth}}{\textbf{Ex 1} ($w_{con}$: 0.64) | \textbf{P:} Because marginal costs are very low, a newspaper price for preprints might be as low as 5 or 6 cents per piece. \textbf{H:} Newspaper preprints can cost as much as \$5.} \\
\midrule
\textbf{Q2B-n} \newline
Qwen3.5-2B \newline \textit{non-thinking} & 
Newspaper preprints can cost as much as \$5 because marginal costs are very low, allowing for a price point that is both affordable and sustainable.  \newline \textit{(32 tokens)} & 
\textcolor{green!70!black}{\checkmark} / \textcolor{red}{$\times$} / \textcolor{red}{$\times$} \\
\midrule

\textbf{Q2B-t} \newline
Qwen3.5-2B \newline \textit{thinking} \newline & 
\textit{Reasoning trace:} \textit{1. Analyze the Request:[...] 2. Analyze the Contradiction: […] 3. Find the Reconciliation [...] \hlc{blue!10}{Likely explanation: There is a difference between the unit price and total cost.} Perhaps there is a subscription, shipping fee, bulk discount? [...] Alternative interpretation: Maybe the \$5 figure refers to a premium edition or shipping included. [...] Wait, I must check if there are any hidden tricks. [...]} \newline \newline
Although the per-unit cost is minimal, buying a large number of copies means the overall expense reaches \$5. \newline \textit{(4984 tokens)} & 
\textcolor{green!70!black}{\checkmark} / \textcolor{green!70!black}{\checkmark} / \textcolor{green!70!black}{\checkmark} \\
\midrule

\textbf{5.6Sol-n} \newline
GPT-5.6-sol \newline \textit{non-thinking} & 
\hlc{blue!10}{The low price applies to individual copies}, while a bundle of roughly 100 preprints could cost about \$5. \newline \textit{(27 tokens)} & 
\textcolor{green!70!black}{\checkmark} / \textcolor{green!70!black}{\checkmark} / \textcolor{green!70!black}{\checkmark} \\
\midrule

\multicolumn{3}{p{0.95\linewidth}}{\textbf{Ex 2} ($w_{con}$: 0.79) | \textbf{P:} What Ellison is doing here, as Hemingway did, is equating the process of becoming an artist with that of becoming a man. \textbf{H:} Ellison and Hemingway took different ways to compare becoming a man.} \\
\midrule
\textbf{G2B-n} \newline
Gemma-4-E2B-it \newline \textit{non-thinking} & 
Ellison's focus on artistic creation reflects a specific facet of the broader human journey that Hemingway addressed through his own literary explorations. \newline \textit{(27 tokens)} & 
\textcolor{red}{$\times$} / \textcolor{green!70!black}{\checkmark} / \textcolor{red}{$\times$} \\
\midrule
\textbf{G26B-n} \newline
Gemma-26B-A4B \newline \textit{non-thinking} & 
While they share the same underlying theme, they employ distinct literary techniques and metaphors to draw this comparison. \newline \textit{(21 tokens)} & 
\textcolor{green!70!black}{\checkmark} / \textcolor{green!70!black}{\checkmark} / \textcolor{green!70!black}{\checkmark} \\
\midrule

\textbf{5.6Sol-t} \newline
GPT-5.6-sol \newline \textit{thinking} & 
\textit{Reasoning trace: [...] I’m working on reconciling how both the artist and the man are compared, but they do so in different ways. [...] Something like, “Although both make this equation, Ellison develops it through...,” and then for Hemingway, [...]} \newline \newline
Although both draw this parallel, Ellison approaches it through racial identity while Hemingway emphasizes courage and self-reliance. \newline \textit{(139 tokens)} & 
\textcolor{green!70!black}{\checkmark} / \textcolor{green!70!black}{\checkmark} / \textcolor{green!70!black}{\checkmark} \\

\midrule

\multicolumn{3}{p{0.95\linewidth}}{\textbf{Ex 3} ($w_{con}$: 0.7) | \textbf{P:} The village is Sainte-Marie, named by the explorer when he landed on 4 November 1493, attracted by the waterfalls and river he could see flowing down the green inland mountains. \textbf{H:} The village is not named after the settling explorer.} \\
\midrule
\textbf{G2B-t} \newline
Gemma-4-E2B-it \newline \textit{thinking} & 
\textit{Reasoning trace: [...] 3. Identify the Contradiction: Premise suggests: Explorer named it (implied meaning). Hypothesis states: \hlc{red!10}{Explorer did not name it.} [...] Brainstorm Reconciliation Explanations \hlc{red!10}{(How can a place be named after someone without the person having done the naming?)}: - It was named by locals. - The name refers to a feature/mythology associated with the explorer. [...] Draft Explanations [...] Attempt 2 (Focusing on coincidence): Although the explorer landed there, \hlc{red!10}{the villagers adopted the name Sainte-Marie} because of a local religious tradition tied to the powerful waters. (Better, focuses on local adoption.) [...]} \newline \newline
\hlc{red!10}{The name Sainte-Marie was bestowed by local inhabitants}, referencing the abundance of water and the lush mountain landscape observed by the newcomer. \newline \textit{(567 tokens)} & 
\textcolor{green!70!black}{\checkmark} / \textcolor{red}{$\times$} / \textcolor{red}{$\times$} \\

\bottomrule
\end{tabular}
\caption{Examples of model-generated explanations in our main evaluation. Color-highlighted extracts are discussed in Appendix \ref{app:qualitative_analysis}. $w_{con}$ refers to the proportion of human annotators that had labeled the premise (P)-hypothesis (H) pair as a contradiction in \citet{nie-etal-2020-learn}. Token counts displayed combine both reasoning (where model is in thinking mode) and final output tokens. E: Efficiency, C: Coherence, S: Success, as judged by Gemma-4-31B-it.}
\label{tab:qual_analysis_p1}
\end{table*}

\begin{table*}[t]
\footnotesize
\setlength{\tabcolsep}{4pt}
\begin{tabular}{p{0.16\linewidth} p{0.77\linewidth} c}
\toprule
\textbf{Model} & \textbf{Explanation} & \textbf{E / C / S} \\
\midrule

\multicolumn{3}{p{0.95\linewidth}}{\textbf{Ex 4} ($w_{con}$: 0.95) | \textbf{P:} All of them slept in one cave on animal skins, a single large clay pot cooked all of their food. \newline \textbf{H:} They used multiple clay pots to cook their food.} \\
\midrule
\textbf{5.6Sol-t} \newline
GPT-5.6-sol \newline \textit{thinking} & 
\textit{Reasoning trace: I need to clarify [...] premise mentions a single large pot cooking all food, but I wonder if multiple pots could have been used over time. [...]} \newline \newline
\hlc{blue!10}{The single cooking pot was replaced several times as it wore out}, so multiple clay pots were used over time.  \newline \textit{(184 tokens) }& 
\textcolor{green!70!black}{\checkmark} / \textcolor{green!70!black}{\checkmark} / \textcolor{green!70!black}{\checkmark} \\
\midrule

\textbf{Q9B-n} \newline
Qwen3.5-9B \newline \textit{non-thinking} & 
The single large clay pot mentioned was likely a communal vessel shared by the group, whereas individual pots were not used for cooking. \newline \textit{(26 tokens)} & 
\textcolor{red}{$\times$} / \textcolor{green!70!black}{\checkmark} / \textcolor{red}{$\times$} \\
\midrule

\textbf{Q9B-t} \newline
Qwen3.5-9B \newline \textit{thinking} & 
\textit{Reasoning trace: [...] 2. Identify the Conflict: - Premise: 1 pot. - Hypothesis: Multiple pots. [...] 3. Brainstorm Reconciliations: - \hlc{blue!10}{Idea 1: They started with one but broke it and got more?} [...] Better Angle: [...]the hypothesis could refer to a broader collection of vessels they possessed [...] Let's stick with the temporal expansion. Draft: Their resourcefulness allowed them to replace the sole vessel with a set of multiple clay pots over time. [...]} \newline \newline
Despite the initial reliance on a solitary vessel, the group eventually secured additional pots to manage their growing culinary requirements.  \textit{\newline (5251 tokens) }& 
\textcolor{green!70!black}{\checkmark} / \textcolor{green!70!black}{\checkmark} / \textcolor{green!70!black}{\checkmark} \\

\bottomrule
\end{tabular}
\caption{Examples of LLM-generated explanations (continued from Table \ref{tab:qual_analysis_p1}).}
\label{tab:qual_analysis_p2}
\end{table*}

\section{Sampling Parameters}\label{app:sampling_params}

As recommended by \cite{qwen3-5}, When prompting models in non-thinking mode, we set temperature at 0.7, \textit{top p} at 0.8, \textit{top k} at 20, \textit{min p} at 0.0, and use a presence penalty of 1.5 and a repetition penalty of 1.0. We use these same sampling parameters when prompting models in thinking mode, except for temperature which we set at 1.0 and \textit{top p} which we set at 0.95. 

In our evaluation, we chose not to use self-consistency \cite{wang2023selfconsistency} (or similar multi-sample decoding approaches) as our task is one of open-ended generation which, unlike tasks such as MCQs, lacks a discrete and well-bounded output space that allows a consensus to be clearly and easily identified. In our case, identifying a genuine consensus among diverse sampled outputs would have required sophisticated semantic clustering that risks overlooking subtle linguistic nuances.

\section{Comparing Explanation Assessment by Humans versus LLM Judge}\label{app:comparing_humans_v_llm_judge}

We obtain 104 samples for annotation by sampling 4 explanations from each of the 13 models in our evaluation prompted in non-thinking and thinking mode ($4 \times 13 \times 2 = 104$). We ensure that each explanation corresponds to a unique premise-hypothesis instance that Gemma-4-31B-it had labeled as ``\textit{Contradiction}''. 

For each premise-hypothesis-explanation triple, each annotator was asked (with instructions to respond with either ``\textit{yes}'' or ``\textit{no}''):
\begin{itemize}
    \item Q1. Is there an apparent contradiction between the Premise and the Hypothesis?
    \item Q2. Is the Explanation coherent and plausible given the Premise?
    \item Q3. Is the Explanation, if true, effective in either  (a) reconciling the apparent contradiction between the Premise and the Hypothesis (if your response to Q1 is yes); or (b) demonstrating that no such apparent contradiction exists (if your response to Q1 is no)?

\end{itemize}

Specifically, Q2 is phrased to correspond to our task criterion of explanation coherence, and Q3 to our criterion of explanation effectiveness. On this basis, we interpret an annotator as considering an explanation successful if their responses to Q2 and Q3 are both ``\textit{yes}''. This mirrors our main evaluation setup as, by our task definition, an LLM judge considers the explanation successful if it considers the explanation both coherent and effective.

Then, we evaluate inter-annotator agreement and the alignment of our LLM judge on the basis of explanation success. 

We find that the two human annotators agreed on 75.96\% of the 104 examples, with a Cohen’s kappa $\kappa$ of 0.427 \cite{Cohen1960}. The LLM judge achieved 76.92\% agreement with Human 1 ($\kappa=0.392$) and 72.12\% agreement with Human 2 ($\kappa=0.374$). These kappa values indicate moderate agreement between the LLM judge and each human annotator, comparable to the human–human agreement. Fleiss’ kappa \cite{fleiss1971measuring} among three annotators (both humans plus LLM judge) is 0.391.

To assess the LLM judge’s alignment with human consensus, we examine the subset of 79 examples for which both human annotators assigned the same success label. Treat this agreed label as the gold standard, we find that the LLM judge correctly classified 65 of the 79 examples as successful, yielding an accuracy of 82.28\%. Additionally, we find a precision of 0.571, recall of 0.706, F1 of 0.632 and a Matthews correlation coefficient (MCC) \cite{MATTHEWS1975442} of 0.522 indicating a positive association between the LLM’s predictions and the human consensus labels.

Taken altogether, we interpret these results as supporting that the LLM judge achieves a level of agreement with human annotators comparable to the agreement observed between the annotators themselves.

\section{Qualitative analysis}\label{app:qualitative_analysis}

Tables \ref{tab:qual_analysis_p1} and \ref{tab:qual_analysis_p2} present examples of model-generated explanations, as discussed below.

When prompted in non-thinking mode, we observe instances of small models simply reusing phrases from the premise and hypothesis without adding any meaningful explanatory content (Ex.1-Q2B-n), and generating seemingly fluent text that however does not address the premise-hypothesis contradiction (Ex.2-G2B-n), both resulting in unsuccessful explanations. 

While these failure modes can be mitigated by prompting the model in thinking mode (compare Ex.1-Q2B-n and Ex.1-Q2B-t i.e. without and with thinking), we also observe cases of \hlc{red!10}{error propagation}: for example, in Ex.3-G2B, the model misparaphrased the hypothesis early in the reasoning chain, and this mistake affected its subsequent generation of candidate explanations, resulting in failure. 

As illustrated by Ex.1-Q2B-t and Ex.4-Q9B-t, Qwen 3.5 models exhibit certain overthinking tendencies. In these examples, at around 6\% and 7\% into the generation process respectively, the model has already come up with \hlc{blue!10}{candidate explanations which we consider likely to succeed} (since they are semantically identical to \hlc{blue!10}{successful explanations} generated by other models: see Ex.1-5.6Sol-n and Ex.4-5.6Sol-t). Yet the model continues to generate tokens redrafting the explanation, considering other ideas and repeatedly checking against the premise and hypothesis.

\section{Additional Results with gpt-5.6-sol as Judge}\label{app:different_judge_exp}

FIrst, we apply bootstrapping (\textit{n=1000}) to results of our main experiment in Section \ref{sec:main_results} and display the resulting 95\% confidence intervals in Table \ref{tab:explanation_metrics_full_w_bootstrap}. 

We then replicate our main experiment using the same setup as detailed Section \ref{sec:experiment} but with gpt-5.6-sol instead of Gemma-4-31B-it acting as NLI judge. As in our main experiment, we filter for the subset of premise-hypothesis pairs in ChaosNLI-MNLI-C that gpt-5.6-sol had considered a contradiction, resulting in 206 instances.

As shown in Table~\ref{tab:explanation_metrics_gpt_sol}, our general findings remain consistent across the two judge models: overall explanation success remains mediocre, ranging from 10.19 (Qwen3.5-0.8B, non-thinking) to 61.65 (gpt-5.6-sol, thinking). 

Compared to Gemma-4-31B-it, gpt-5.6-sol assigns lower coherence scores for explanations generated by all 13 models (in both non-thinking and thinking modes). Notably, 9 of the 10 coherence scores for the Qwen 3.5 model family are below the corresponding 95\% lower confidence intervals of our baseline results. In contrast, effectiveness scores increase with respect to 22 out of 26 model-mode combinations.

Despite these differences, explanation success rates as judged by the two different models remain broadly consistent: only two of the 26 scores fall outside the 95\% confidence intervals of our baseline results (Qwen3.5-9B and gpt-5.6-luna, both in thinking mode). This relative stability demonstrates the robustness of explanation success as our primary task criterion across different judge models.

Finally, when judging explanations generated by Gemma-4-31B-it, gpt-5.6-sol assigns an explanation success rate of 46.12 in non-thinking mode and 49.03 in thinking mode, respectively 0.48 and 1.85 higher than when Gemma-4-31B-it is judging its own explanations in our baseline results. As such, whilst we recognize concerns raised in existing work about the risk of models exhibiting self-preferential bias when acting as judge (e.g. \citealp{wataoka2024selfpreference}; \citealp{xu-etal-2024-pride}; \citealp{10.5555/3737916.3740113}), we identified no evidence of such bias with respect to Gemma-4-31B-it acting as judge for our specific task.

\begin{table*}[t]
\centering
\footnotesize
\renewcommand{\arraystretch}{1}
\setlength{\tabcolsep}{2pt}
\newcommand{\scoreci}[3]{\shortstack[c]{#1\\[-0.35ex]{\scriptsize [#2, #3]}}}
\begin{tabular}{p{1.3cm} l c c c c c c}
\hline
& \multirow{2}{*}{\textbf{\shortstack{Source Model\\of Generated\\Explanations}}} & \multicolumn{2}{c}{\textbf{Coherence}} & \multicolumn{2}{c}{\textbf{Effectiveness}} & \multicolumn{2}{c}{\textbf{Explanation Success}} \\
\cmidrule(lr){3-4} \cmidrule(lr){5-6} \cmidrule(lr){7-8}
& & \textbf{\shortstack{Non-\\Thinking}} & \textbf{Thinking} & \textbf{\shortstack{Non-\\Thinking}} & \textbf{Thinking} & \textbf{\shortstack{Non-\\Thinking}} & \textbf{Thinking} \\
\hline
\multirow{5}{=}{\textit{\textcolor{gray}{Qwen 3.5}}}
& Qwen3.5-0.8B & \scoreci{72.31}{66.15}{78.46} & \scoreci{72.82}{66.65}{79.49} & \scoreci{26.67}{21.03}{32.82} & \scoreci{22.05}{16.41}{27.18} & \scoreci{11.28}{7.18}{15.90} & \scoreci{16.41}{11.28}{21.03} \\
& Qwen3.5-2B & \scoreci{66.15}{59.49}{72.32} & \scoreci{66.15}{59.49}{73.33} & \scoreci{38.46}{31.79}{45.13} & \scoreci{57.95}{50.77}{64.10} & \scoreci{18.46}{13.33}{24.10} & \scoreci{33.33}{26.67}{39.49} \\
& Qwen3.5-4B & \scoreci{76.41}{70.77}{81.54} & \scoreci{74.36}{68.21}{80.51} & \scoreci{17.95}{12.82}{23.08} & \scoreci{57.44}{50.77}{64.12} & \scoreci{13.33}{8.72}{17.95} & \scoreci{40.51}{33.85}{47.69} \\
& Qwen3.5-9B & \scoreci{86.67}{82.05}{91.28} & \scoreci{81.54}{75.90}{86.67} & \scoreci{30.26}{24.10}{37.44} & \scoreci{64.10}{57.44}{70.77} & \scoreci{24.62}{18.46}{30.77} & \scoreci{51.28}{44.62}{57.95} \\
& Qwen3.5-27B & \scoreci{75.90}{69.74}{81.54} & \scoreci{80.51}{74.87}{86.15} & \scoreci{61.54}{54.36}{68.21} & \scoreci{70.77}{63.58}{76.94} & \scoreci{44.10}{36.92}{51.28} & \scoreci{54.36}{47.18}{61.03} \\
\hline
\multirow{5}{=}{\textit{\textcolor{gray}{Gemma 4}}}
& Gemma-4-E2B-it & \scoreci{87.18}{82.56}{91.79} & \scoreci{82.05}{76.92}{86.67} & \scoreci{26.67}{20.51}{33.33} & \scoreci{30.77}{24.10}{37.95} & \scoreci{20.00}{14.36}{25.64} & \scoreci{23.08}{17.42}{29.23} \\
& Gemma-4-E4B-it & \scoreci{\textbf{90.77}}{86.67}{94.36} & \scoreci{85.13}{80.00}{90.26} & \scoreci{18.46}{13.33}{24.62} & \scoreci{38.46}{31.28}{45.14} & \scoreci{13.33}{8.72}{17.96} & \scoreci{31.28}{24.62}{37.95} \\
& Gemma-4-12B-it & \scoreci{82.56}{76.92}{87.69} & \scoreci{80.51}{74.87}{85.64} & \scoreci{43.59}{36.91}{50.77} & \scoreci{57.95}{50.77}{64.62} & \scoreci{30.77}{24.62}{37.44} & \scoreci{46.67}{39.49}{53.33} \\
& Gemma-4-26B-A4B-it & \scoreci{80.51}{74.36}{86.15} & \scoreci{85.13}{80.51}{90.26} & \scoreci{47.69}{41.03}{54.87} & \scoreci{55.38}{48.21}{62.56} & \scoreci{34.36}{27.69}{41.04} & \scoreci{44.10}{37.42}{50.77} \\
& \fcolorbox{blue}{white}{Gemma-4-31B-it} & \scoreci{81.54}{75.90}{86.67} & \scoreci{81.03}{75.38}{86.67} & \scoreci{57.44}{50.26}{64.12} & \scoreci{62.05}{54.87}{68.72} & \scoreci{45.64}{38.46}{52.31} & \scoreci{47.18}{40.00}{54.36} \\
\hline\hline
\multirow{3}{=}{\textit{\textcolor{gray}{OpenAI\\(propri-\\etary)}}}
& gpt-5.6-luna & \scoreci{81.54}{75.38}{86.67} & \scoreci{\textbf{89.23}}{85.13}{93.33} & \scoreci{49.23}{42.05}{56.41} & \scoreci{70.26}{64.10}{76.92} & \scoreci{35.38}{28.72}{42.05} & \scoreci{62.05}{55.38}{69.23} \\
& gpt-5.6-terra & \scoreci{84.10}{78.46}{89.23} & \scoreci{81.03}{75.38}{86.15} & \scoreci{55.90}{48.72}{62.56} & \scoreci{77.44}{71.28}{83.59} & \scoreci{44.10}{36.92}{50.77} & \scoreci{61.03}{53.85}{67.18} \\
& gpt-5.6-sol & \scoreci{85.64}{80.00}{90.26} & \scoreci{80.00}{74.36}{85.13} & \scoreci{\textbf{73.33}}{66.67}{79.49} & \scoreci{\textbf{79.49}}{73.33}{85.13} & \scoreci{\textbf{62.05}}{54.87}{68.72} & \scoreci{\textbf{62.56}}{55.90}{69.74} \\
\hline\hline
\end{tabular}
\caption{Extended table containing our main results in Table \ref{tab:explanation_metrics_full}, with 95\% bootstrap confidence intervals enclosed in brackets. As before, coherence, effectiveness, and explanation success rate (in \%) of each model's generated explanations are as evaluated by Gemma-4-31B-it as the judge model (enclosed in \fcolorbox{blue}{white}{blue}). Scores in the row corresponding to Gemma-4-31B-it are when the model acts as both explanation generator and judge. Highest value in each column is in \textbf{bold}.}
\label{tab:explanation_metrics_full_w_bootstrap}
\end{table*}

\begin{table*}[t]
\centering
\footnotesize
\renewcommand{\arraystretch}{1.2}
\setlength{\tabcolsep}{2pt}
\begin{tabular}{p{1.3cm} l c c c c c c}
\hline
& \multirow{2}{*}{\textbf{\shortstack{Source Model\\of Generated\\Explanations}}} & \multicolumn{2}{c}{\textbf{Coherence}} & \multicolumn{2}{c}{\textbf{Effectiveness}} & \multicolumn{2}{c}{\textbf{Explanation Success}} \\
\cmidrule(lr){3-4} \cmidrule(lr){5-6} \cmidrule(lr){7-8}
& & \textbf{\shortstack{Non-\\Thinking}} & \textbf{Thinking} & \textbf{\shortstack{Non-\\Thinking}} & \textbf{Thinking} & \textbf{\shortstack{Non-\\Thinking}} & \textbf{Thinking} \\
\hline
\multirow{5}{=}{\textit{\textcolor{gray}{Qwen 3.5}}} 
& Qwen3.5-0.8B & \shortstack{\fcolorbox{red}{white}{60.68} \\ {\tiny\textcolor{red}{(-11.63)}}} & \shortstack{\fcolorbox{red}{white}{60.19} \\ {\tiny\textcolor{red}{(-12.63)}}} & \shortstack{26.21 \\ {\tiny\textcolor{red}{(-0.46)}}} & \shortstack{\fcolorbox{red}{white}{32.04} \\ {\tiny\textcolor{green!70!black}{(+9.99)}}} & \shortstack{10.19 \\ {\tiny\textcolor{red}{(-1.09)}}} & \shortstack{16.02 \\ {\tiny\textcolor{red}{(-0.39)}}} \\
& Qwen3.5-2B   & \shortstack{\fcolorbox{red}{white}{57.77} \\ {\tiny\textcolor{red}{(-8.38)}}} & \shortstack{\fcolorbox{red}{white}{57.77} \\ {\tiny\textcolor{red}{(-8.38)}}} & \shortstack{43.69 \\ {\tiny\textcolor{green!70!black}{(+5.23)}}} & \shortstack{64.08 \\ {\tiny\textcolor{green!70!black}{(+6.13)}}} & \shortstack{18.45 \\ {\tiny\textcolor{red}{(-0.01)}}} & \shortstack{31.55 \\ {\tiny\textcolor{red}{(-1.78)}}} \\
& Qwen3.5-4B   & \shortstack{\fcolorbox{red}{white}{67.96} \\ {\tiny\textcolor{red}{(-8.45)}}} & \shortstack{\fcolorbox{red}{white}{66.50} \\ {\tiny\textcolor{red}{(-7.86)}}} & \shortstack{\fcolorbox{red}{white}{25.24} \\ {\tiny\textcolor{green!70!black}{(+7.29)}}} & \shortstack{62.14 \\ {\tiny\textcolor{green!70!black}{(+4.70)}}} & \shortstack{14.08 \\ {\tiny\textcolor{green!70!black}{(+0.75)}}} & \shortstack{38.35 \\ {\tiny\textcolor{red}{(-2.16)}}} \\
& Qwen3.5-9B   & \shortstack{\fcolorbox{red}{white}{79.61} \\ {\tiny\textcolor{red}{(-7.06)}}} & \shortstack{\fcolorbox{red}{white}{67.96} \\ {\tiny\textcolor{red}{(-13.58)}}} & \shortstack{34.95 \\ {\tiny\textcolor{green!70!black}{(+4.69)}}} & \shortstack{63.59 \\ {\tiny\textcolor{red}{(-0.51)}}} & \shortstack{23.30 \\ {\tiny\textcolor{red}{(-1.32)}}} & \shortstack{\fcolorbox{red}{white}{41.75} \\ {\tiny\textcolor{red}{(-9.53)}}} \\
& Qwen3.5-27B  & \shortstack{70.39 \\ {\tiny\textcolor{red}{(-5.51)}}} & \shortstack{\fcolorbox{red}{white}{69.90} \\ {\tiny\textcolor{red}{(-10.61)}}} & \shortstack{67.96 \\ {\tiny\textcolor{green!70!black}{(+6.42)}}} & \shortstack{76.21 \\ {\tiny\textcolor{green!70!black}{(+5.44)}}} & \shortstack{45.15 \\ {\tiny\textcolor{green!70!black}{(+1.05)}}} & \shortstack{51.94 \\ {\tiny\textcolor{red}{(-2.42)}}} \\
\hline
\multirow{5}{=}{\textit{\textcolor{gray}{Gemma 4}}} 
& Gemma-4-E2B-it     & \shortstack{\fcolorbox{red}{white}{77.67} \\ {\tiny\textcolor{red}{(-9.51)}}} & \shortstack{\fcolorbox{red}{white}{73.30} \\ {\tiny\textcolor{red}{(-8.75)}}} & \shortstack{28.64 \\ {\tiny\textcolor{green!70!black}{(+1.97)}}} & \shortstack{\fcolorbox{red}{white}{39.32} \\ {\tiny\textcolor{green!70!black}{(+8.55)}}} & \shortstack{17.48 \\ {\tiny\textcolor{red}{(-2.52)}}} & \shortstack{25.24 \\ {\tiny\textcolor{green!70!black}{(+2.16)}}} \\
& Gemma-4-E4B-it     & \shortstack{\fcolorbox{red}{white}{\textbf{85.92}} \\ {\tiny\textcolor{red}{(-4.85)}}} & \shortstack{\textbf{81.55} \\ {\tiny\textcolor{red}{(-3.58)}}} & \shortstack{19.90 \\ {\tiny\textcolor{green!70!black}{(+1.44)}}} & \shortstack{40.78 \\ {\tiny\textcolor{green!70!black}{(+2.32)}}} & \shortstack{16.50 \\ {\tiny\textcolor{green!70!black}{(+3.17)}}} & \shortstack{32.04 \\ {\tiny\textcolor{green!70!black}{(+0.76)}}} \\
& Gemma-4-12B-it     & \shortstack{79.13 \\ {\tiny\textcolor{red}{(-3.43)}}} & \shortstack{76.70 \\ {\tiny\textcolor{red}{(-3.81)}}} & \shortstack{44.17 \\ {\tiny\textcolor{green!70!black}{(+0.58)}}} & \shortstack{55.83 \\ {\tiny\textcolor{red}{(-2.12)}}} & \shortstack{32.04 \\ {\tiny\textcolor{green!70!black}{(+1.27)}}} & \shortstack{39.81 \\ {\tiny\textcolor{red}{(-6.86)}}} \\
& Gemma-4-26B-A4B-it & \shortstack{79.13 \\ {\tiny\textcolor{red}{(-1.38)}}} & \shortstack{80.58 \\ {\tiny\textcolor{red}{(-4.55)}}} & \shortstack{45.63 \\ {\tiny\textcolor{red}{(-2.06)}}} & \shortstack{56.80 \\ {\tiny\textcolor{green!70!black}{(+1.42)}}} & \shortstack{33.01 \\ {\tiny\textcolor{red}{(-1.35)}}} & \shortstack{42.23 \\ {\tiny\textcolor{red}{(-1.87)}}} \\
& Gemma-4-31B-it     & \shortstack{80.58 \\ {\tiny\textcolor{red}{(-0.96)}}} & \shortstack{76.70 \\ {\tiny\textcolor{red}{(-4.33)}}} & \shortstack{58.25 \\ {\tiny\textcolor{green!70!black}{(+0.81)}}} & \shortstack{65.05 \\ {\tiny\textcolor{green!70!black}{(+3.00)}}} & \shortstack{46.12 \\ {\tiny\textcolor{green!70!black}{(+0.48)}}} & \shortstack{49.03 \\ {\tiny\textcolor{green!70!black}{(+1.85)}}} \\
\hline\hline
\multirow{3}{=}{\textit{\textcolor{gray}{OpenAI\\(propri-\\etary)}}} 
& gpt-5.6-luna  & \shortstack{75.73 \\ {\tiny\textcolor{red}{(-5.81)}}} & \shortstack{\fcolorbox{red}{white}{75.73} \\ {\tiny\textcolor{red}{(-13.50)}}} & \shortstack{52.43 \\ {\tiny\textcolor{green!70!black}{(+3.20)}}} & \shortstack{75.73 \\ {\tiny\textcolor{green!70!black}{(+5.47)}}} & \shortstack{32.04 \\ {\tiny\textcolor{red}{(-3.34)}}} & \shortstack{\fcolorbox{red}{white}{55.34} \\ {\tiny\textcolor{red}{(-6.71)}}} \\[1.5ex]
& gpt-5.6-terra & \shortstack{\fcolorbox{red}{white}{77.18} \\ {\tiny\textcolor{red}{(-6.92)}}} & \shortstack{76.21 \\ {\tiny\textcolor{red}{(-4.82)}}} & \shortstack{61.17 \\ {\tiny\textcolor{green!70!black}{(+5.27)}}} & \shortstack{82.04 \\ {\tiny\textcolor{green!70!black}{(+4.60)}}} & \shortstack{42.23 \\ {\tiny\textcolor{red}{(-1.87)}}} & \shortstack{60.68 \\ {\tiny\textcolor{red}{(-0.35)}}} \\[1.5ex]
& \fcolorbox{blue}{white}{gpt-5.6-sol} & \shortstack{80.10 \\ {\tiny\textcolor{red}{(-5.54)}}} & \shortstack{75.24 \\ {\tiny\textcolor{red}{(-4.76)}}} & \shortstack{\textbf{73.79} \\ {\tiny\textcolor{green!70!black}{(+0.46)}}} & \shortstack{\textbf{83.01} \\ {\tiny\textcolor{green!70!black}{(+3.52)}}} & \shortstack{\textbf{59.71} \\ {\tiny\textcolor{red}{(-2.34)}}} & \shortstack{\textbf{61.65} \\ {\tiny\textcolor{red}{(-0.91)}}} \\[1.5ex]
\hline\hline
\end{tabular}
\caption{Coherence, effectiveness, and success rate of each model's generated explanations (in \%), as evaluated by \textbf{gpt-5.6-sol} as the judge model (enclosed in \fcolorbox{blue}{white}{blue}). Displayed delta figures (\textcolor{green!70!black}{+}/\textcolor{red}{-}) are computed against baseline results which use Gemma-4-31B-it as judge. Values outside the 95\% bootstrap confidence intervals of the baseline results (see Table \ref{tab:explanation_metrics_full_w_bootstrap}) are enclosed in \fcolorbox{red}{white}{red}. Scores in the row corresponding to gpt-5.6-sol are when the model itself acts as both explanation generator and the judge doing the evaluating. Highest score in each column is in \textbf{bold}.}
\label{tab:explanation_metrics_gpt_sol}
\end{table*}

\section{Datasets Used and Licenses}

\begin{center}
\renewcommand{\arraystretch}{1.2}

\begin{tabular}{ll}

\hline
\textbf{Dataset} & \textbf{License} \\ \hline 
\shortstack{ChaosNLI\\ \cite{nie-etal-2020-learn}}         & CC BY-NC 4.0     
\\
\hline
\end{tabular}
\end{center}

As described in Section \ref{subsec:dataset}, given the diverse sources of English sentences constituting the ChaosNLI-MNLI subset used in our study (e.g. travel guides and news reports), some of these generic sentences contain the names of historical and public figures. 

We manually inspect samples from ChaosNLI-MNLI \cite{nie-etal-2020-learn} to check that they contain no offensive content.

\section{Hyperparameters Used in Training Experiment}\label{app:training_hyperparams}

As described in Section \ref{sec:model_training}, we use the default hyperparameters specified by the GRPOTrainer and GRPOConfig classes in the TRL library, varying only the upper clipping threshold (\textit{$epsilon\_high$}).\footnote{See \url{https://huggingface.co/docs/trl/v1.15.0/en/grpo_trainer\#trl.GRPOConfig}} Notably, these include training for 3 epochs, with \textit{beta} at 0.0 (i.e. we follow the approach adopted by DAPO's authors in not penalizing for KL divergence \cite{yu2025dapo}), and lower clipping threshold at 0.2. 

During the rollout phase, the model generates its outputs in non-thinking mode, using the same sampling parameters as we had used in our main evaluation experiment for non-thinking mode generation, again detailed in Appendix \ref{app:sampling_params}.

\end{document}